\documentclass[submission,copyright,creativecommons]{eptcs}
\providecommand{\event}{ICLP 2020} 
\usepackage{underscore}           

\usepackage{xspace}
\usepackage{multirow}
\usepackage{graphics}
\usepackage{graphicx}  
\usepackage{lscape}
\usepackage{url}
\usepackage{epsfig}
\usepackage{longtable}
\usepackage{color}
\usepackage{amsfonts}
\usepackage{amsmath}
\usepackage{xspace}
\usepackage{epsfig}

\usepackage[]{algorithm2e}
\usepackage{caption}
\usepackage{subcaption}
\usepackage{float}
\usepackage{dblfloatfix}
\usepackage{here}

\def\mapf{{\bf MAPF}\xspace}

\def\dmapf{{\bf D-MAPF}\xspace}

\def\mapf{{\bf MAPF}\xspace}

\def\dmapf{{\bf D-MAPF}\xspace}

\def\clingo{{\sc Clingo}\xspace}
\def\lar{\leftarrow}
\def\ba{\begin{array}}
\def\ea{\end{array}}
\def\beq{\begin{equation}}
\def\eeq#1{\label{#1}\end{equation}}

\def\ii#1{\hbox{\it #1\/}}

\def\conflictSet{{\it conflictSet}\xspace}

\def\asp{{\sc ASP}\xspace}

\title{Dynamic Multi-Agent Path Finding based on\\ Conflict Resolution using Answer Set Programming}
\author{ Basem Atiq and Volkan Patoglu and Esra Erdem
\institute{Faculty of Engineering and Natural Sciences, Sabanci University, Istanbul, Turkey}
\email{ \{basem, vpatoglu, esra.erdem\}@sabanciuniv.edu}
}
\def\titlerunning{Dynamic MAPF based on Conflict Resolution using ASP}
\def\authorrunning{B. Atiq, V. Patoglu, E. Erdem}
\begin{document}
\maketitle

\begin{abstract}
We study a dynamic version of multi-agent path finding problem (called \dmapf) where existing agents may leave and new agents may join the team at different times. We introduce a new method to solve \dmapf based on conflict-resolution. The idea is, when a set of new agents joins the team and there are conflicts, instead of replanning for the whole team, to replan only for a minimal subset of agents whose plans conflict with each other. We utilize answer set programming as part of our method for planning, replanning and identifying minimal set of conflicts.
\end{abstract}

\section{Introduction}

Multi-agent path finding (\mapf) problem aims to find paths for multiple agents from their initial locations to destinations such that no two agents collide with each other while they follow these paths. This problem has been studied under various constraints (e.g., where an upper bound is given on the plan length) or attempting to reach a certain objective (e.g., minimizing the total time taken for all agents to reach their goals, or minimizing the maximum time taken for each agent to reach its goal location). All these variants are NP-hard~\cite{RatnerW86,Surynek10}.

We study a dynamic version of the \mapf problem that emerges when changes in our environment begin to take place, e.g., when new agents are added to the team at different times with their own initial and goal locations, or when some obstacles are placed into the environment. We refer to this problem as Dynamic Multi-Agent Path Finding (\dmapf) problem. \dmapf\ has many direct applications in automated  warehouses, where teams of hundreds of robots are utilized to prepare dynamic orders in an every changing environment~\cite{Wurman}.

We propose a new method to solve the \dmapf problem, which involves replanning for a small set of agents that conflict with each other. When several new agents join the team, if some conflicts occur, our objective is to minimize the number of agents that are required to replan to resolve these conflicts. In this way, we avoid having to replan for all agents and rather, keep the plans of as many of the existing agents fixed. We identify a minimal set of agents whose paths should be replanned by means of identifying conflicts and then resolving them by replanning.

The proposed method utilizes Answer Set Programming (ASP)~\cite{MarekT99,Niemelae99,Lifschitz02} (based on answer sets~\cite{GelfondL88,GelfondL91}) for planning, replanning and identifying a minimal set of agents with conflicts. The ASP formulation used for planning is presented in our earlier study~\cite{ErdemKOS13}, to which we refer the reader for details. In the following, we will focus more on the use of ASP for the latter two problems.

\begin{figure}[t]
	\centering
   \resizebox{\columnwidth}{!}{\includegraphics{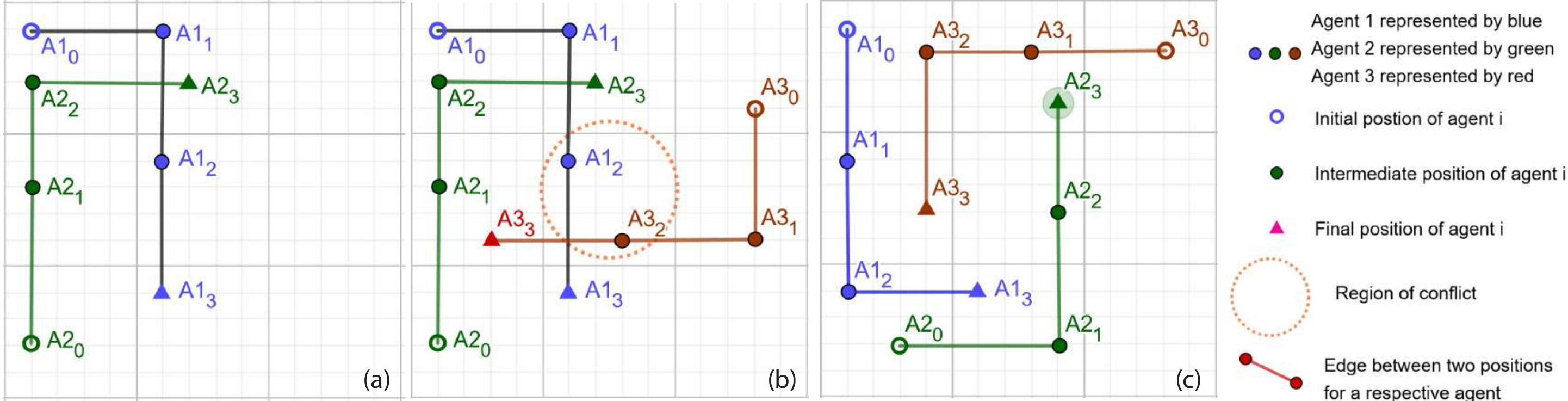}}
    \vspace{-1.\baselineskip}
    \caption{An illustrative example. (a) 2 agents with their already determined respective paths. (b) A new agent $a_3$ is added to the environment but cannot find a collision-free path. (c) All agents replan their solutions to find a collision free path. }
	\label{fig:sh1}
\end{figure}

\section{Dynamic \mapf}

\dmapf can be thought of as a generalization of the \mapf problem. In the case of \dmapf, we deal with changes that take place with the passage of time. These changes can include, but are not be limited to, the addition of obstacles into the environment, the addition of new agents into the environment, and the changes in the objectives of each agent for a given problem.

The inputs to a \dmapf problem are the same as that for the \mapf problem which includes the initial and goal positions of each agent, the updates or modifications that have taken place in the environment, a restriction on the makespan of each agent, and the paths of existing agents. Figure~\ref{fig:sh1} above gives an example of a \dmapf problem where a new agent $a_3$ is added to the existing environment which consists of two agents $a_1$ and $a_2$ who already have their paths determined as in Figure~\ref{fig:sh1}(a). With a makespan of each agent restricted to 3, agent $a_3$ is unable to find a collision free solution as shown in Figure~\ref{fig:sh1}(b) for the given instance.

As the main objective of any \mapf problem is to find a collision free solution for all agents, replanning is attempted for all agents in the example as shown in Figure~\ref{fig:sh1}(c).

\section{Solving \dmapf via Conflict Resolution}
\label{sec:conflict}

We introduce a new method to solve \dmapf where replanning for all agents is avoided most of the time. This method keeps track of two sets of agents throughout the program: \textit{nonConflictSet}, which contains the set of agents (and their plans) that do not conflict with each other and, ideally, remain as they are despite the changes in the environment; and \textit{conflictSet}, which contains the set of agents (and their plans) that conflict with each other, and, ideally, replanning for a minimal subset of this set would resolve conflicts.

Our algorithm applies when some new agents join the team, as the existing agents are executing their plans.

\begin{enumerate}
\item When a set of new agents join the set of existing agents, then try to find a \mapf solution for the new agents so that they do not conflict with each other or the existing agents.
\item If such a solution exists, then include the new agents (with their plans) in \textit{nonConflictSet}.
\item Otherwise, include the new agents (with their plans) in \textit{conflictSet}.
\item While there is some conflict to resolve do the following:
\begin{enumerate}
\item Try to find a minimal(-cardinality) subset of agents in \textit{conflictSet}, such that replanning for them resolves the conflicts in \textit{conflictSet}.
\item If such a minimal subset of agents is found, then include all agents (and their plans/replans) from the \textit{conflictSet} into \textit{nonConflictSet}.
\item Otherwise, some conflicts exist between some agents in \textit{nonConflictSet} and \textit{conflictSet}, expand \textit{conflictSet} by a set of agents (and their plans) from \textit{nonConflictSet} that cause the minimum number of conflicts.
\item Meanwhile, move the agents from \textit{conflictSet} that are not involved in these conflicts to \textit{nonConflictSet}.
\end{enumerate}
\end{enumerate}

Note that, in the worst case, the algorithm above replans for all agents.

We use a slight variation of the ASP formulation $\Pi$ for \mapf from our earlier studies~\cite{ErdemKOS13} to find a \mapf solution for the new agents in Step 1 above, by generating plans for the new agents only and by incorporating the plans of the existing agents as facts.

In Step 4(a), we enumerate all subsets of \textit{conflictSet} with cardinality 2,3,... incrementally, and use a slight variation of $\Pi$ for replanning for each subset of agents only and by incorporating the plans of the other agents in \textit{conflictSet} as facts.

In Step 4(c), we expand the \textit{conflictSet} by utilizing \asp's noteworthy feature of weak constraints. In particular, we identify the minimum number of conflicts between agents in the \textit{conflictSet} and those in the \textit{nonConflictSet}:
$$
\xleftarrow{\scriptstyle\sim} \ii{plan}(t,a_1,x,y),\ \ii{path}(t,a_2,x,y),\ \ii{conflictSet}(a_1), \ \ii{nonConflictSet}(a_2)\ [1@1,a_1,a_2,t] \qquad (a_1\neq a_2).
$$
Here, a penalty of 1 is assigned each time such a conflict is detected.  ASP solver generates several solutions with the addition of this weak constraint, however, a solution with the lowest penalty cost is chosen. In addition to the weak constraints, note that we still include hard constraints to prevent collisions between agents within the conflict set:
$$
\ba l
\lar \ii{path}(t,a_1,x,y),\ \ii{path}(t,a_2,x,y),\ \ii{conflictSet}(a_1),\ \ii{conflictSet}(a_2) \quad(a_1\neq a_2) .
\ea
$$

Figure~\ref{fig:sh2} below gives an example of a scenario where our algorithm manages to find a collision free solution for the agents in the environment without having to replan for all agents. The existing agents ($a_1, a_2, a_3, a_4$) are added to the \textit{nonConflictSet} and their existing paths are stored. The three new agents ($a_5, a_6, a_7$) are added to the environment.

The algorithm first attempts to find a solution for the new agents while keeping the paths of the pre-existing agents fixed. Unable to find a solution, it places the new agents in \textit{conflictSet}. It further tries to resolve conflicts within \textit{conflictSet}. Unable to resolve the conflicts, the algorithm tries to expand the conflict set by trying to find a minimum set of conflicts between agents in \textit{conflictSet} and the agents in \textit{nonConflictSet}. The algorithm finds, as shown in Figure~\ref{fig:sh2}(c), that the agents $a_1, a_2, a_4, a_5$ and $a_6$ conflict amongst each other. Then, \textit{conflictSet} is updated to contain these agents only, while $a_7$ is moved to \textit{nonConflictSet}.

\begin{figure}[t]
	\centering
    \resizebox{\columnwidth}{!}{\includegraphics{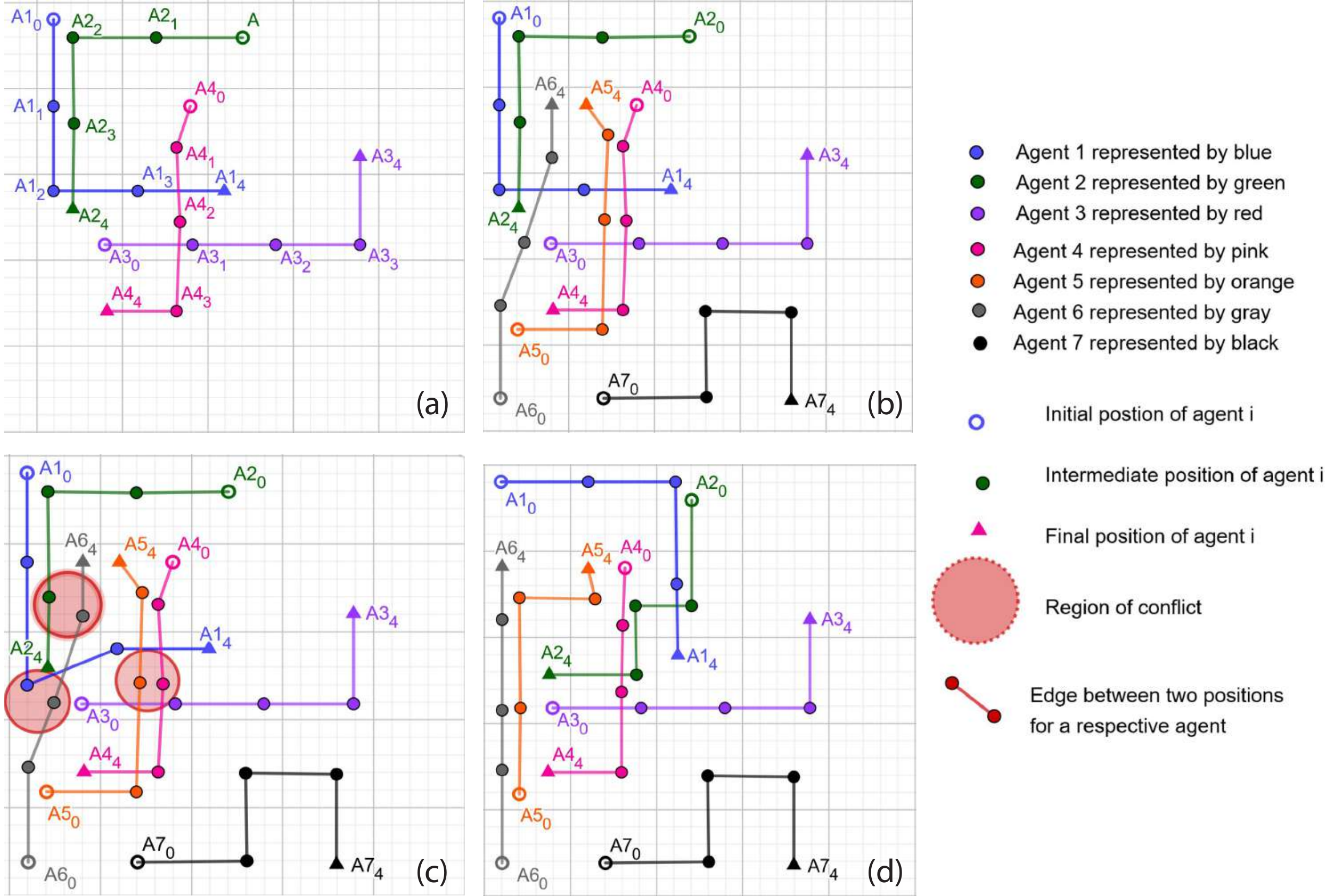}}
    \vspace{-1.2\baselineskip}
    \caption{An illustrative example. (a)~4~agents with their already determined respective paths  are added to the \textit{nonConflictSet}. (b)~3~new agents $a_5$, $a_6$, $a_7$ are added to the \textit{conflictSet} and  a solution is attempted. (c)~Regions of conflicts amongst agents in the \textit{conflictSet} and the \textit{nonConflictSet} are shown. (d)~New plans for all agents are presented.  }
	\label{fig:sh2} \vspace{-.25\baselineskip}
\end{figure}

The algorithm then proceeds to resolve the conflicts within $\textit{conflictSet}= \{a_1, a_2, a_4, a_5, a_6\}$. It enumerates all subsets of size $\geq 2$ of the agents in the \textit{conflictSet}:  $\{a_1,a_2\}, \{a_1,a_4\}, \{a_1,a_5\}, \{a_1,a_6\}, \{a_2,a_4\}$, $\{a_2,a_5\}, \{a_2,a_6\}, \{a_4,a_5\}, \{a_4,a_6\}, \{a_5,a_6\}$.
Each subset is selected one at a time, and the algorithm proceeds to determine whether a solution can be found by replanning only for those two agents in the given subset. In this particular case, the algorithm is unable to find a solution for any of the 10 subsets of size 2. Then the algorithm enumerates all subsets of size 3: $\{a_1,a_2,a_4\}, \{a_1,a_2,a_5\}, \{a_1,a_2,a_6\}, \{a_1,a_4,a_5\}$, $\{a_1,a_4,a_6\}, \{a_1,a_5,a_6\}, \{a_2,a_4,a_5\}, \{a_2,a_4,a_6\}, \{a_2,a_5,a_6\}, \{a_4,a_5,a_6\}$. Once again the algorithm attempts to replan for each subset one at a time until a solution is found. Fortunately, this time the algorithm is able to find a solution for the subset $\{a_1,a_2,a_5\}$ and replanning is performed only for those agents to devise a collision free solution for all agents as shown in Figure~\ref{fig:sh2}(d).

\section{Experimental Evaluations} \vspace{-1mm}

We have compared our algorithm to solve \dmapf with the straightforward approach of replanning, by means of some experiments. The algorithm described in the previous section has been implemented using Python~3.6.4 and \clingo~4.5.4, and we have performed experiments on a Linux server with 16~2.4~GHz~Intel~E5-2665 CPU cores and 64~GB memory.

\begin{table}[t!]
\centering
\caption{Experimental evaluations.}
\vspace{-.5\baselineskip}
\resizebox{1.0\columnwidth}{!}{\begin{tabular}{|c|c|c|c|c|c|c|c|c|c|}
\hline
Initial  & \# of new & Makespan    & \multicolumn{2}{c}{Replanning for a Subset}    & \multicolumn{2}{c}{Replanning for All} & Cardinality of the & Cardinality of the subset & Subset \\
Instance &  agents   &  & CPU time [s] & Solution Found [Y/N] & CPU time [s] & Solution Found [Y/N] & Conflict Set & For which solution is found &  Number\\
\hline \hline
 & 1 & 38 & 1.90  & Y & 49.40   & Y & 2 & 2 & 1  \\
 1 & 2 & 38 & 4.58  & Y & 50.58   & Y & 4 & 2 & 3 \\
 28 agents & 3 & 38 & 15.16  & Y & 53.49   & Y & 6 & 3 & 16   \\
 $20\times 20$ grid & 4 & 38 & 142.66  & Y & 56.57  & Y & 8 & 4 & 120  \\
 \hline
 & 1 & 58 & 7.34  & Y & 180.63   & Y & 2 & 2 & 1  \\
 2 & 2 & 58 & 18.07  & Y & 186.77   & Y & 4 & 2 & 3 \\
 28 agents & 3 & 58 & 57.90  & Y & 247.68  & Y & 6 & 3 & 16  \\
 $30\times 30$ grid & 4 & 58 & 551.74  & Y & 261.07  & Y & 8 & 4 & 120  \\
 \hline
 & 1 & 78 & 21.83  & Y & $>1000$ & Y & 2 & 2 & 1  \\
 3 & 2 & 78 & 50.28  & Y & $>1000$ & Y & 4 & 2 & 3   \\
 42 agents & 3 & 78 & 61.72  & Y & $>1000$ & Y & 4 & 2 & 3 \\
 $40\times 40$ grid & 4 & 78 & 62.59  & Y & 1440.14  & Y & 5 & 2 & 1  \\
  \hline
 & 1 & 98 & 39.67 & Y & $>2500$   & Y & 2 & 2 & 1  \\
 4 & 2 & 98 & 115.74  & Y & $>2500$   & Y & 4 & 2 & 4  \\
 38 agents & 3 & 98 & 172.40  & Y & $>2500$   & Y & 4 & 2 & 4 \\
 $50\times 50$ grid & 4 & 98 & 394.36 & Y & 2766.37  & Y & 6 & 3 & 16 \\
 \hline
 5 & & & & & & & & & \\
 46 agents & 4 & 138 & 23408.82 & Y & -  & N & 8 & 4 & 85 \\
 $70\times 70$ grid & & & & & & & & & \\
 \hline
\end{tabular}}\label{Tab:results} 
\end{table}

Experiments have been carried for various grid sizes and varying number of agents as shown in Table~\ref{Tab:results}. For each grid size, 4 test cases were run. The number of existing agents and the makespan for a particular grid size have been kept fixed for each of the 4 test cases. For uniformity, the makespan has been selected as the longest path from one corner of the grid to the opposite diagonal. The tests have also been carried out with the assumption that no static obstacles exist and that each agent starts at $t=0$. The size of the \conflictSet along with the size of the subset for which a solution is found is also shown for better analysis.

To serve as an example, let us look at the forth instance in Table~\ref{Tab:results} with 38 agents on a $50\times 50$ grid, with a makespan of 98. When four new agents are added to the environment, a new solution is computed by our algorithm in 394.36 seconds whereas replanning for all agents requires 2766.37 seconds. The number of agents that were conflicting with each other in this case were 6 and the size of the subset for which a solution was found was 3.

For small grid sizes, replanning for all agents outperformed our implementation. This was expected, however, as calling the \asp program with weak constraints generates many more possible configurations and for such small instances, it is more efficient to replan for all agents. The results get more interesting as the grid size and the number of agents increase. When the grid size increases, we obtain the results as expected in almost all of the remaining test cases.

There were exceptions to the efficiency of our algorithm as shown by the last test case for $20 \times 20$ grid sizes. Replanning for all agents proved to be more efficient than our version because 120 subsets had to be tried until a solution was found. This proved to be a more time consuming process, therefore the results are as indicated.

Results for the $40 \times 40$ and $50 \times 50$ instances show how much more effective it can be to replan only for a subset of agents. For the test case with a grid size of $40 \times 40$ and 4 new agents, our algorithm was at least 20 times as efficient as replanning for all agents.

As the grid size and the number of agents increases further, there is a notable difference between the time taken to find a solution by our algorithm and replanning. For the largest grid size of $70 \times 70$, our algorithm  found a solution in about 6 hours. However, replanning for all agents  was not possible due to the sheer size of the input.

From these results, we observe that the underlying idea of reusing existing solutions may be quite efficient in terms of computation time, in particular, for large instances with a large makespan.

\section{Related Work}

{\em Regarding conflicts}: A sort of conflict-resolution has been utilized by the Conflict-Based Search Algorithm (CBS)~\cite{SharonSFS15} introduced to solve the \mapf problem. The approach attempts to decompose a \mapf problem into several constrained single-agent path finding problems. At the high level, the algorithm maintains a binary tree referred to as a Conflict Tree (CT) which detects conflicts and adds a set of constraints for every agent to each node of the tree. At the low level, the shortest path for every agent with respect to its constraints are searched for. The algorithm then checks to determine whether any conflicts arise with the new paths computed at that node. If conflicts do arise, the algorithm declares the current node as a non-goal node. What is interesting about their approach is the way that they deal with conflicts. While we generate all possible subsets of the conflicting agents, attempt to replan for each subset until a solution is found or expand the conflict set, \cite{SharonSFS15} splits the node at which a conflict arises into its two children nodes and both these nodes are then checked to see if a solution exists. If a conflict exists between two agents $\alpha_i$ and $\alpha_j$, each child node contains an additional constraint to its parent node for either $\alpha_i$ or $\alpha_j$. Search is then performed for only the agent which is associated with the new constraint while the paths of all other agents are kept fixed. When conflicts are generated amongst more than 2 agents, focus is placed on the first two agents and the same procedure as described above is followed. Further conflicts are dealt with at a deeper level of the tree.

{\em Regarding dynamic \mapf}: Online \mapf~\cite{svancara2019} considers the addition of new agents to the team while a plan is being executed, under the assumptions that agents disappear when they reach their goal and that new agents may wait before entering their initial location in the environment. These assumptions relax the \dmapf problem: the new agents may enter the environment one at a time, and they provide more space for the other agents when they disappear. To solve online \mapf with these assumptions, Svancara et al. investigate algorithms that rely on replanning (e.g., for all agents) and conflict-resolution (e.g., planning for the new agents one at a time ignoring others, and then resolving conflicts by replanning). Our approach does not rely on such assumptions and tries to resolve conflicts by identifying the minimal set of agents that cause conflicts.

In an earlier study~\cite{BogatarkanP019}, we introduce an alternative method for \dmapf. It does not rely on conflict-resolution. The idea is to revise the traversals of paths of the existing agents (up to the given upper bound on makespan) while computing new plans for the new agents so that there is no conflict between any two agents. If a solution cannot be found, then replanning is applied for all agents. We plan to use the method based on conflict-resolution, in combination with the revise and augment method to further reduce the number of replannings as part of our ongoing studies.

\section{Conclusion}

Our approach to minimizing the number of agents that are required to replan their solutions has been shown to be very efficient as detailed above. Replanning for all agents tends to become expensive very quickly once our environment becomes larger or more congested. An alternate approach as described by our algorithm can help reduce the cost of performing such a search while minimizing the modifications applied to the paths of agents that already exist.

\section*{Acknowledgements}

This work has been partially supported by Tubitak Grant 188E931.

\bibliographystyle{eptcs}

\begin{thebibliography}{10}
\providecommand{\bibitemdeclare}[2]{}
\providecommand{\surnamestart}{}
\providecommand{\surnameend}{}
\providecommand{\urlprefix}{Available at }
\providecommand{\url}[1]{\texttt{#1}}
\providecommand{\href}[2]{\texttt{#2}}
\providecommand{\urlalt}[2]{\href{#1}{#2}}
\providecommand{\doi}[1]{doi:\urlalt{http://dx.doi.org/#1}{#1}}
\providecommand{\bibinfo}[2]{#2}

\bibitemdeclare{inproceedings}{BogatarkanP019}
\bibitem{BogatarkanP019}
\bibinfo{author}{Aysu \surnamestart Bogatarkan\surnameend},
  \bibinfo{author}{Volkan \surnamestart Patoglu\surnameend} \&
  \bibinfo{author}{Esra \surnamestart Erdem\surnameend} (\bibinfo{year}{2019}):
  \emph{\bibinfo{title}{A Declarative Method for Dynamic Multi-Agent Path
  Finding}}.
\newblock In: {\sl \bibinfo{booktitle}{Proc. of GCAI}}, \bibinfo{volume}{65},
  pp. \bibinfo{pages}{54--67}, \doi{10.1609/aaai.v33i01.33017732}.

\bibitemdeclare{inproceedings}{ErdemKOS13}
\bibitem{ErdemKOS13}
\bibinfo{author}{Esra \surnamestart Erdem\surnameend},
  \bibinfo{author}{Doga~Gizem \surnamestart Kisa\surnameend},
  \bibinfo{author}{Umut \surnamestart Oztok\surnameend} \&
  \bibinfo{author}{Peter \surnamestart Schueller\surnameend}
  (\bibinfo{year}{2013}): \emph{\bibinfo{title}{A General Formal Framework for
  Pathfinding Problems with Multiple Agents}}.
\newblock In: {\sl \bibinfo{booktitle}{Proc. of AAAI}}.
\newblock \bibinfo{note}{\url{https://dl.acm.org/doi/10.5555/2891460.2891501}}.

\bibitemdeclare{inproceedings}{GelfondL88}
\bibitem{GelfondL88}
\bibinfo{author}{M.~\surnamestart Gelfond\surnameend} \&
  \bibinfo{author}{V.~\surnamestart Lifschitz\surnameend}
  (\bibinfo{year}{1988}): \emph{\bibinfo{title}{The stable model semantics for
  logic programming}}.
\newblock In: {\sl \bibinfo{booktitle}{Proc. of ICLP}}, \bibinfo{publisher}{MIT
  Press}, pp. \bibinfo{pages}{1070--1080}.

\bibitemdeclare{article}{GelfondL91}
\bibitem{GelfondL91}
\bibinfo{author}{Michael \surnamestart Gelfond\surnameend} \&
  \bibinfo{author}{Vladimir \surnamestart Lifschitz\surnameend}
  (\bibinfo{year}{1991}): \emph{\bibinfo{title}{Classical negation in logic
  programs and disjunctive databases}}.
\newblock {\sl \bibinfo{journal}{New Generation Computing}}
  \bibinfo{volume}{9}, pp. \bibinfo{pages}{365--385}, \doi{10.1007/bf03037169}.

\bibitemdeclare{article}{Lifschitz02}
\bibitem{Lifschitz02}
\bibinfo{author}{Vladimir \surnamestart Lifschitz\surnameend}
  (\bibinfo{year}{2002}): \emph{\bibinfo{title}{Answer set programming and plan
  generation}}.
\newblock {\sl \bibinfo{journal}{Artificial Intelligence}}
  \bibinfo{volume}{138}, pp. \bibinfo{pages}{39--54},
  \doi{10.1016/s0004-3702(02)00186-8}.

\bibitemdeclare{incollection}{MarekT99}
\bibitem{MarekT99}
\bibinfo{author}{Victor \surnamestart Marek\surnameend} \&
  \bibinfo{author}{Miros{\l}aw \surnamestart Truszczy\'nski\surnameend}
  (\bibinfo{year}{1999}): \emph{\bibinfo{title}{Stable models and an
  alternative logic programming paradigm}}.
\newblock In: {\sl \bibinfo{booktitle}{The Logic Programming Paradigm: a
  25-Year Perspective}}, \bibinfo{publisher}{Springer Verlag}, pp.
  \bibinfo{pages}{375--398}, \doi{10.1007/978-3-642-60085-2_17}.

\bibitemdeclare{article}{Niemelae99}
\bibitem{Niemelae99}
\bibinfo{author}{Ilkka \surnamestart Niemel{\"a}\surnameend}
  (\bibinfo{year}{1999}): \emph{\bibinfo{title}{Logic programs with stable
  model semantics as a constraint programming paradigm}}.
\newblock {\sl \bibinfo{journal}{Annals of Mathematics and Artificial
  Intelligence}} \bibinfo{volume}{25}, pp. \bibinfo{pages}{241--273},
  \doi{10.1023/A:1018930122475}.

\bibitemdeclare{inproceedings}{RatnerW86}
\bibitem{RatnerW86}
\bibinfo{author}{Daniel \surnamestart Ratner\surnameend} \&
  \bibinfo{author}{Manfred~K. \surnamestart Warmuth\surnameend}
  (\bibinfo{year}{1986}): \emph{\bibinfo{title}{Finding a Shortest Solution for
  the N $\times$ N Extension of the 15-PUZZLE Is Intractable}}.
\newblock In: {\sl \bibinfo{booktitle}{Proc. of AAAI}}, pp.
  \bibinfo{pages}{168--172}.
\newblock \bibinfo{note}{\url{https://dl.acm.org/doi/10.5555/2887770.2887797}}.

\bibitemdeclare{article}{SharonSFS15}
\bibitem{SharonSFS15}
\bibinfo{author}{Guni \surnamestart Sharon\surnameend}, \bibinfo{author}{Roni
  \surnamestart Stern\surnameend}, \bibinfo{author}{Ariel \surnamestart
  Felner\surnameend} \& \bibinfo{author}{Nathan~R. \surnamestart
  Sturtevant\surnameend} (\bibinfo{year}{2015}):
  \emph{\bibinfo{title}{Conflict-based search for optimal multi-agent
  pathfinding}}.
\newblock {\sl \bibinfo{journal}{Artif. Intell.}} \bibinfo{volume}{219}, pp.
  \bibinfo{pages}{40--66}, \doi{10.1016/j.artint.2014.11.006}.

\bibitemdeclare{inproceedings}{Surynek10}
\bibitem{Surynek10}
\bibinfo{author}{Pavel \surnamestart Surynek\surnameend}
  (\bibinfo{year}{2010}): \emph{\bibinfo{title}{An Optimization Variant of
  Multi-Robot Path Planning Is Intractable}}.
\newblock In: {\sl \bibinfo{booktitle}{Proc. of AAAI}}.
\newblock \bibinfo{note}{\url{https://dl.acm.org/doi/10.5555/2898607.2898808}}.

\bibitemdeclare{inproceedings}{svancara2019}
\bibitem{svancara2019}
\bibinfo{author}{Jiri \surnamestart Svancara\surnameend},
  \bibinfo{author}{Marek \surnamestart Vlk\surnameend}, \bibinfo{author}{Roni
  \surnamestart Stern\surnameend}, \bibinfo{author}{Dor \surnamestart
  Atzmon\surnameend} \& \bibinfo{author}{Roman \surnamestart Bartak\surnameend}
  (\bibinfo{year}{2019}): \emph{\bibinfo{title}{Online Multi-Agent
  Pathfinding}}.
\newblock In: {\sl \bibinfo{booktitle}{Proc. of {AAAI}}},
  \doi{10.1609/aaai.v33i01.33017732}.

\bibitemdeclare{article}{Wurman}
\bibitem{Wurman}
\bibinfo{author}{Peter \surnamestart Wurman\surnameend},
  \bibinfo{author}{Raffaello \surnamestart D'Andrea\surnameend} \&
  \bibinfo{author}{Mick \surnamestart Mountz\surnameend}
  (\bibinfo{year}{2008}): \emph{\bibinfo{title}{Coordinating Hundreds of
  Cooperative, Autonomous Vehicles in Warehouses}}.
\newblock {\sl \bibinfo{journal}{AI Magazine}} \bibinfo{volume}{29}, pp.
  \bibinfo{pages}{9--20}, \doi{10.1609/aimag.v29i1.2082}.

\end{thebibliography}

\end{document}